\documentclass{bmvc2k}

\title{Learning Filter Scale and Orientation In CNNs}

\addauthor{Ilker Cam}{ilker.cam@isikun.edu.tr}{1}
\addauthor{F. Boray Tek}{boray.tek@isikun.edu.tr}{1}

\addinstitution{
 Computer Engineering\\
 Isik University\\
 Istanbul, TR
}

\runninghead{Cam \& Tek}{Learning Filter Scale and Orientation In CNNs}

\usepackage{graphicx,subfigure}
\usepackage{amsmath,amsfonts,amsthm,bm} 

\graphicspath{{images/}}
\DeclareGraphicsExtensions{.pdf,.jpeg,.png, .svg}

\begin{document}

\maketitle

\begin{abstract}
	
Convolutional neural networks have many hyperparameters such as the filter size, number of filters, and pooling size, which require manual tuning. Though deep stacked structures are able to create multi-scale and hierarchical representations, manually fixed filter sizes limit the scale of representations that can be learned in a single convolutional layer. 

This paper introduces a new adaptive filter model that allows variable scale and orientation. The scale and orientation parameters of filters can be learned using back propagation. Therefore, in a single convolution layer, we can create filters of different scale and orientation that can adapt to small or large features and objects. The proposed model uses a relatively large base size (grid) for filters. In the grid, a differentiable function acts as an envelope for the filters. The envelope function guides effective filter scale and shape/orientation by masking the filter weights before the convolution. Therefore, only the weights in the envelope are updated during training.

In this work, we employed a multivariate (2D) Gaussian as the envelope function and showed that it can grow, shrink, or rotate by updating its covariance matrix during back propagation training . We tested the new filter model on MNIST, MNIST-cluttered, and CIFAR-10 and compared the results with the networks that used conventional convolution layers. The results demonstrate that the new model can effectively learn and produce filters of different scales and orientations in a single layer. Moreover, the experiments show that the adaptive convolution layers perform equally; or better, especially when data includes objects of varying scale and noisy backgrounds.

\end{abstract}

\section{Introduction}

Naming or describing real life objects is only meaningful with respect to a relevant scale \cite{Lindeberg:1994:STC:528688}. For example, a view can be described as a leaf, a branch, or a tree depending on the distance of the observer. Natural and casual scenes are generally composed of many different entities/objects at different scales. During image acquisition, the true physical scale is usually ignored. However, the relative scale of the objects is somehow implicitly captured and stored in the image grid and pixels.

An automated method to identify or describe objects in images can be analyzed in two parts: representation + classification. Basic classification algorithms without add-ons can not successfully handle variation and complexity of raw pixel-level representation of objects, instead they rely on functions that map image pixels into different constructs, -named features-, which are sought to represent the image content more briefly and invariant to various geometric and intensity changes.

Traditionally, computer vision researchers relied on manually designed feature extractors for representation. Recently, we are witnessing the success of the algorithms which can self-learn appropriate feature extractors. In either case, the size of an operator or a probe usually determines and fixes scale of the entities that can be represented. However, even in the self-learn case, size of the probes or operators is often manually selected. On the other hand, last two decades has seen many automated object detection/recognition algorithms that were superior to their counterparts because they have comprised multi-scale processing of images \cite{dalal2005histograms}, \cite{lenet}. Multi-scale feature extractors gather and present the inherent scale information of image pixels to a subsequent classifier. In SIFT \cite{lowe} and wavelets \cite{mallat1992characterization}, this is done by creating a multi-scale pyramid from the input image and then applying a fixed size probe-kernel to each scale. In an application of Gabor filters for object recognition, Serre et al. \cite{serre} used a hierarchy of stacked Gabor filtering layers, where the filters have predetermined scales and orientations. However, Chan et al. \cite{pcanet} showed that the adaptation of handcrafted filters to low-level representations is difficult. On the other hand, convolution neural networks (CNN) rely on stacked and hierarchical convolutions of the image to extract multi-scale information. Convention of CNNs for filter size selection is to use small fixed size weight kernels in the lower levels. However, thanks to the stacked operation of convolutional layers, sandwiched by pooling layers which down-sample intermediate outputs, the deeper levels of a network are able to learn representations of larger scales. Though the optimality of fixed size kernels has not proven, the convention is to use filters small as 3x3 in the first layer, which can be larger 5x5 or 7x7 in the later stages \cite{Zeiler2014}. During back propagation training, filters are evolved to imitate lower level receptive fields in biological vision which are sensitive to certain shapes and orientations. Another justification for avoiding large filter sizes is that, while certainly increasing computation time, they may also increase over-fitting.

Though the number of filters and their sizes in convolution layers are usually selected intuitively, researchers are seeking alternatives to improve representation capacity of the network in deeper architectures. For example, Szegedy et al. \cite{inceptionv1} handcrafted their ``inception'' architecture to include mixing of parallel and wide convolution layers which use different sized filter kernels. In a deep architecture, this approach allows multi-scale, parallel and sparse representations.
\cite{onemoreexample-directionfilters}
In summary, existing CNN based methods use fixed size convolution kernels and then rely on the fact that shape and orientation of the filters can be inferred from the training data. Additionally, CNNs employ stacked convolution layers to successfully create multi-scale representations.


On the other hand, Hubel and Wiesel \cite{hubel1968receptive} discovered three type of cells in visual cortex: simple, complex, hyper-complex (i.e.\ end-stopped cells). The simple cells are sensitive to the orientation of the excitatory input, whereas the hypercomplex cells are activated with a certain orientation, motion, and \emph{length} of the stimuli. Therefore it is biologically plausible to assume that filters of different scales next to different orientations and direction may also work better in CNNs.

In this study, we create a new and adaptive model of convolution layers where the filter size (actually scale) and orientation are learned during training. Therefore, a single convolution layer can have distinct filter scales and orientations. Broadly speaking, this corresponds to extracting multi-scale information using a single convolution layer. However, our aim is not to fully replace the stacked architectures and deep networks for multi-scale information. Instead, our approach improves the information that can be extracted from an input (may be an image), in a single layer. Additionally, the model removes the necessity of fixing convolution kernel sizes, so that the filter size can be removed from the list of hyper-parameters of deep learning networks. 

Our experiments use MNIST, MNIST-cluttered and CIFAR-10 datasets to show that the proposed model can learn and produce different scaled filters in a single convolution layer whilst improving classification performance compared to conventional convolution layers. In experimental side, our work is concentrated on developing and proving an effective methodology for learnable adaptive filter scales and orientations, rather than improving highly optimized state-of-the-art results in these datasets.
Organization of the paper is as follows \ref{}.

\section{Method}

\label{sec:methods}
In deep network architectures, stacked convolution layers perform convolutions with fixed size kernels where sandwiched pooling layers perform downsampling operations to achieve a multi-scale and hierarchal representation. Fixed size convolution kernels put a limit on the scale of features which can be extracted from a single layer. 

\ref{Lets put two figures explaining cnn and our approach.}

Though it is possible to mix several kernels of different size in a single convolution layer, the convention is to use a fixed size for all the kernels in a layer. Here, we introduce a new filter model which can adapt its scale and orientation. Therefore it allows development of multi-scale and differently oriented filters in a single convolution layer. To realize this, we need a smooth and  function that can grow, shrink, or rotate during training which acts as an envelope to guide filter scale and orientation. The following subsections explain the role of the envelope, selecting an appropriate function for the envelope, and its partial derivatives which are used in backpropagation.

\begin{figure}
	\centering     
	
	\subfigure[Arbitrary envelope]{\label{fig1:a}\includegraphics[width=0.23\linewidth]{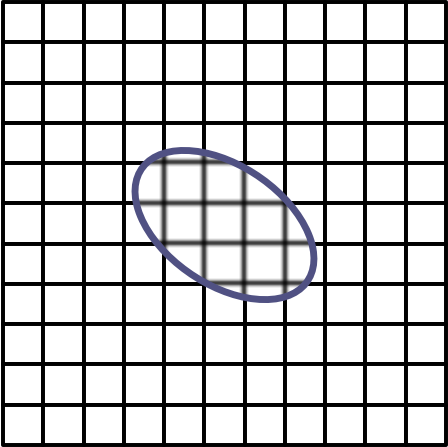}}
	\subfigure[Gaussian envelope]{\label{fig1:b}\includegraphics[width=0.23\linewidth]{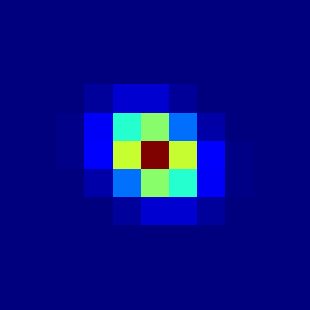}}
	\subfigure[Random weights]{\label{fig1:c}\includegraphics[width=0.23\linewidth]{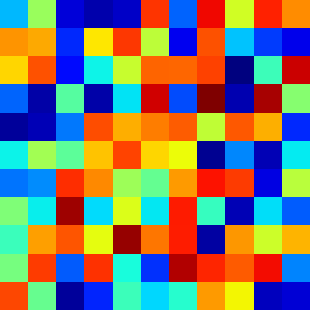}}
	\subfigure[Masked weights]{\label{fig1:d}\includegraphics[width=0.23\linewidth]{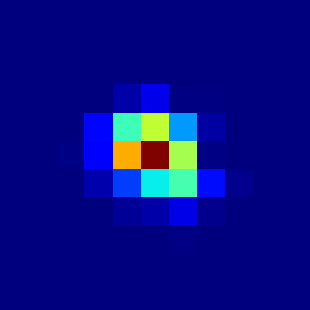}}
	
	\caption{Illustration of the proposed weight envelope (a) an arbitrary differentiable envelope function controls the weight spread and shape on a regular relatively large base grid, (b) an example, initial Gaussian kernel with centered $\mu$ and initial $\Sigma$, (c) initial weights of the filter that are randomly generated, (d) weights are masked with the envelope (b) by simple element-wise multiplication.}
	\label{fig1}
\end{figure}

\subsection{Envelope Function}
\label{sec:methods:envelope}
The role of the envelope function is to guide the filter scale and orientation development. As illustrated in Figure \ref{fig1}, a base grid acts as the envelope and filter domain. Since it is the most common case, we will assume a two-dimensional domain. Generalizations to a higher dimensions is straightforward. In this domain, the envelope function must be differentiable and smooth. Let us assume a base grid for an $n \times n$ odd sized and square filter (\ref{eq:grid}); and let $G$ be a (continuously) differentiable function defined in grid $g$ (coordinate space) and parameter vector $\boldsymbol{\Theta} \in R^i$ to define its shape (\ref{eq:envelope}).

\begin{equation}
\label{eq:grid}
\begin{gathered}
A = \{1, 2, .. , n\},
\\
g \in  A \times A = \{(a, b) \: | \: a \in A \: and \: b \in A\}
\end{gathered}
\end{equation}

\begin{equation}
\label{eq:envelope}
u = U(g, \boldsymbol{\Theta}), \qquad \qquad  \boldsymbol{\Theta} = \{\theta_1, \theta_2, .. \theta_i\}
\end{equation}

By updating the parameters in $\boldsymbol{\Theta}$, envelope function must be able to grow or shrink its effective area and change its orientation. The feed forward model of a single neuron with an input $x$ and transfer function $f$ can be written as: 

\begin{equation}
o = f(\sum\limits_{g \in A \times A} \: x_{g} \: w_{g} \: U({g,\boldsymbol{\Theta}}))
\end{equation}

Or if we think of a whole convolution layer of input matrix (image) $X$ and weight matrix $W$ and envelope matrix $U$. Simply, an element-wise multiplication of $U$ with the weight matrix $W$ will mask and scale the weights before the convolution.

\begin{equation}
O = f(X* (W \circ U))
\end{equation}

Since the weights can not grow out of envelope $U$, the filter size and orientation will be bounded and determined by $U$. Assuming that the partial derivatives of $G$ with respect to continuous parameter $\theta_i$ is defined using the chain rule, the update can be performed using the standard backpropagation algorithm with the learning rate $\eta$. However note that the weight update also gets $u$ as a scaler

\begin{equation}
w_{g}^{'} \mathrel{{:}{=}} w_{g} - \eta \frac{\partial E}{\partial o} \frac{\partial o}{\partial w_{g}}\qquad
\label{eq:genericderive}
\theta_{i}^{'} \mathrel{{:}{=}} \theta_{i} - \eta \frac{\partial E}{\partial o} \frac{\partial o}{\partial U} \frac{\partial U}{\partial w_{g}}
\end{equation}

\subsection{Selecting an Envelope Function}
\label{sec:methods:gaussian}

It is well known that continuous Gaussian kernel has unique properties which are important for generating a scale space. Simply put, the Gaussian kernel does not create new local extrema, nor enhance existing extrema, whilst smoothing the image with a variable continuous parameter \cite{Lindeberg:1994:STC:528688}. Some of this properties are proven to exist in discrete space if the sample size is sufficiently large. Therefore, Gaussian is an ideal candidate for the envelope function $U$:

\begin{equation}
U(g, \mu, \Sigma) = \frac{1}{A} e ^{ - \frac{1}{2} (g-\mu)' \Sigma^{-1} (g-\mu)}
\end{equation}

Here, parameter $A$ is an optional normalization parameter; $\mu$ controls the center of the envelope, whereas the covariance $\Sigma$ controls the scale and orientation of the kernel. 

\begin{equation}
\label{eq:initmu}
\mu = \ \bigg{\{} \mu_x, \mu_y \bigg{\}} \quad
\Sigma =  \begin{bmatrix} \sigma_{xx} & \sigma_{xy} \\ \sigma_{xy} & \sigma_{yy} \end{bmatrix}
\end{equation}

During the feed forward execution the envelope function is calculated on the grid coordinates $g$ with the current covariance $\Sigma$; and then elementwise multiplied with the weight matrix $W$, prior to the convolution. This is illustrated in Figure \ref{fig1:b}-\ref{fig1:d}. Note that this operation not only bounds the weights and adjusts the effective area, it also scales the weights. To implement the convolution operation appropriately, we set $\boldsymbol{\mu}$ as a vector of constants that is initialized with the center point coordinates of the grid $g$. Therefore, it is not updated during training. However, the covariance $\Sigma$ must be updated to learn the filter scale and orientation.



In order to keep the symmetric property of $\Sigma$ we calculate the gradients for each $\sigma$ and apply update rule.


\begin{equation}
\label{eq:derivesigma}
\forall \sigma \in \Sigma, \quad \sigma \mathrel{{:}{=}} \sigma - \eta \frac{\partial E}{\partial U} \frac{\partial U}{\partial \sigma}
\end{equation}

Covariance $\Sigma$ must be kept as a symmetric and positive definite matrix.
A symmetric matrix is positive definite if for all non-zero vectors: $x^T \Sigma x\ge0$; which imposes the following conditions:
\begin{equation}
 \left( \sigma_{xx}>0, \: \: \sigma_{yy}>0, \: \: \sigma_{xx}\sigma_{yy}>\sigma^2_{xy}\right)  \quad
  \mathtt{or} \quad
 \lambda_i \ge 0
\end{equation}

where $\lambda_i$ denotes the eigenvalues of the covariance matrix, which can be checked to ensure positive definiteness. However, during training the diagonal sigma terms are ensured to be positive (and nonzero) by setting bounds, e.g. $\sigma_{xx}=$min$\left(\epsilon,\sigma_{xx}\right)$; whereas $\sigma_{xy}$ is constrained by $\sigma_{xy}$=max$\left(\sigma_{xy},\sqrt{\sigma_{xx}\sigma_{yy}}\right)$ . Experiments show that the covariance behave well during training when it is initialized properly and updated with a small learning rate; and thus it removes the necessity for these constraints. 



\subsection{Vanishing Variance}
\subsection{Implementation}

We implemented the adaptive convolution filters using Lasagne\&\ref{?}Theano \cite{theano} and then tested using a Nvidia Tesla K40 GPU board. In terms of computational complexity, as it can be expected, calculating the Gaussian envelope function adds an extra overhead in training. However, during feed forward execution, the trained and enveloped final weights can be stored and used immediately without any overhead. Compared to conventional filters, we use relatively larger (e.g. 11x11) base filter sizes to observe adaptive growth and rotations. Please note that the grid can be selected as large as the input image. ACNN ran one epoch (500 examples) in $6.1 \pm .2$ seconds whereas Cnn-11 ran in $3.4 \pm .15$ seconds on MNIST dataset without an optimized implementation. (The code will be available in the camera-ready version, not disclosed for anonymity.)

\section{Experiments}
\label{sec:experiments}

\subsection{Why do we need a filter guide?}

One can argue that the filter guide is unnecessary because a relatively large CNN-layer can learn any filter. To disprove this idea we conducted a test with an encoder where the input is an image the filter is expected to learn a simple filtering operation. Can CNN learn the same filter without a filter guide. 

Show learned filter of a Gaussian blur filter with encoder. 

Show learned filter of an edge filter with encoder.

Show learned filter of a blurred edge filter with encoder.

\subsection{Covariance}

Show how covariance learns, whether it requires the constraints, in which conditions. \\

\subsection{Runtime wrt grid size}

\begin{table}
\centering
\caption{The network topology that is used to test our method. All three networks were comprised of 8 layers. In conv-1 and conv-2 layers, the proposed adaptive model (ACNN) used an 11x11 base grid for filters, whereas 'cnn-5' and 'cnn-11' used 5x5 and 11x11 filter sizes, respectively. *CIFAR-10 experiments used 16 filters instead of 8.}
\bigskip
\begin{tabular}{ | l | l | l | l | l | l |}
	\hline
	Layer & Units & Filters & Filter Size & Pool Size & Activation \\ \hline
	1- input & - & - & - & - & - \\ \hline
	2- conv-1 & - & 8/16* & 5x5/11x11 & - & ReLu \\ \hline
	3- maxpool-1 & - & - & - & 2x2 & - \\ \hline
	4- conv-2 & - & 8/16* & 5x5/11x11 & - & ReLu \\ \hline
	5- maxpool-2 & - & - & - & 2x2 & - \\ \hline
	6- dropout(50\%) & - & - & - & - & - \\ \hline
	7- fully connected - 1 & 256 & - & - & - & ReLu \\ \hline
	8- fully connected - 2 & 10 & - & - & - & Sigmoid \\ \hline
\end{tabular}
\label{table:network}

\end{table}

The experiments were aimed at observing whether the adaptive filters can change its scale and orientation during training and whether this adaptation yields an improved classification performance. We tested the adaptive filter model with three different datasets, and also compared the results against two conventional CNN configurations that used different fixed size filters. All three networks had the same structure which was comprised of two convolution, two pooling layers, a dropout layer and two fully connected layers (Table \ref{table:network}). The only difference between the adaptive CNN (ACNN) and conventional CNNs (cnn-5, cnn-11) was the replacement of convolution layers. We used cross-entropy as the error function.
The hyper parameters we used are as follows; Learning rate: 0.01, momentum: 0.95, batch size: 500. On each test, we examined training loss, error \% and scale, orientation changes in filters. Though the learning rate for $\Sigma$ could be adjusted separately it was not necessary.

\subsection{MNIST}

\begin{figure}
	\centering
	\subfigure[Orientation Change.]{\label{fig:mnist:orientation}\includegraphics[width=0.47\linewidth]{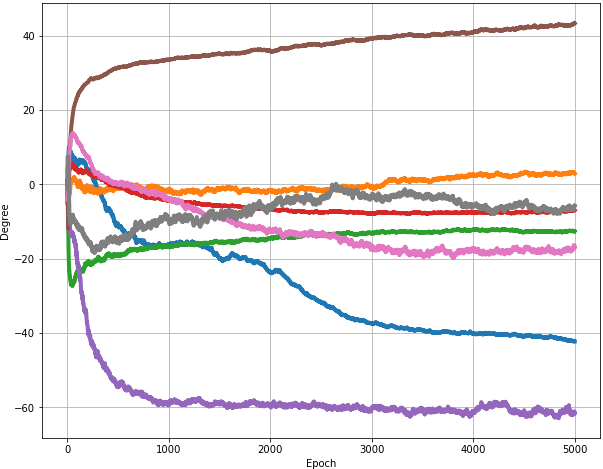}}
	\subfigure[Scale Change.]{\label{fig:mnist:scale}\includegraphics[width=0.47\linewidth]{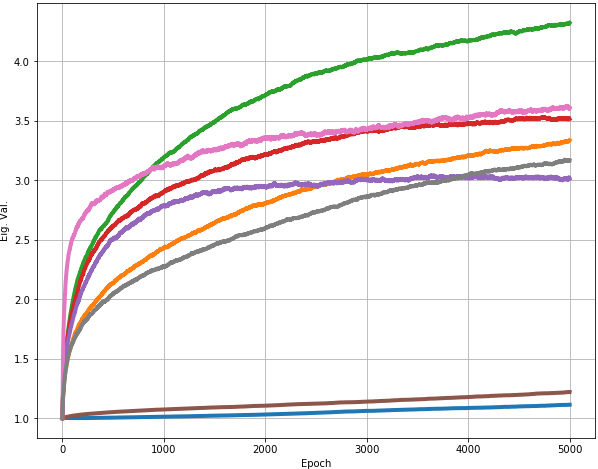}}
	\caption{The plots for covariance matrix $\Sigma$ change in MNIST dataset. Depicted by the (a) angle of largest eigenvector and (b) largest eigenvalue.}
	\label{fig:mnist:convchange}
\end{figure}

MNIST \cite{mnist} is a database of handwritten digits, widely used in machine learning research to test models. It has 50,000 training and 10,000 testing images from 10 different categories. To observe the change in $\Sigma$, we calculated its eigenvalues and eigenvectors. The maximum eigenvalue represents the scale, whereas the tangent between the eigenvectors shows the orientation as illustrated in Figure \ref{fig:mnist:convchange}. 

\begin{figure}
	\centering
	\subfigure[Gaussian envelope functions.]{\label{fig:mnist:filter1:mask}\includegraphics[width=0.32\linewidth]{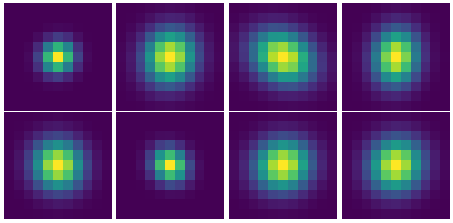}}
	\subfigure[Scaled filters.]{\label{fig:mnist:filter1:scaled}\includegraphics[width=0.32\linewidth]{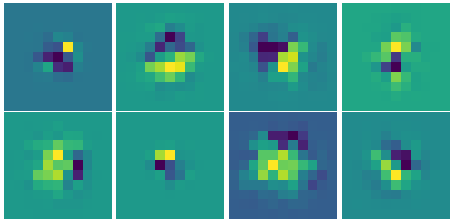}}
	\subfigure[Convolution outputs.]{\label{fig:mnist:filter1:conved}\includegraphics[width=0.32\linewidth]{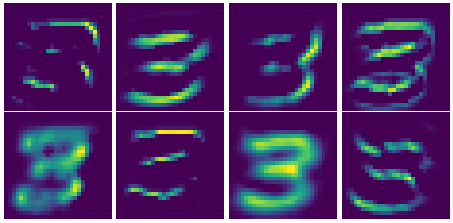}}
	\caption{The first layer (conv-1) filters at the end of training with MNIST. (a) Gaussian envelopes, (b) scaled filters, (c) output of a sample that was convolved with each filter.}
	\label{fig:mnistfilt1}
\end{figure}

\begin{figure}
	\centering 
	\subfigure[Training loss, y axis is on log scale.]{\label{fig:a}\includegraphics[width=0.49\linewidth]{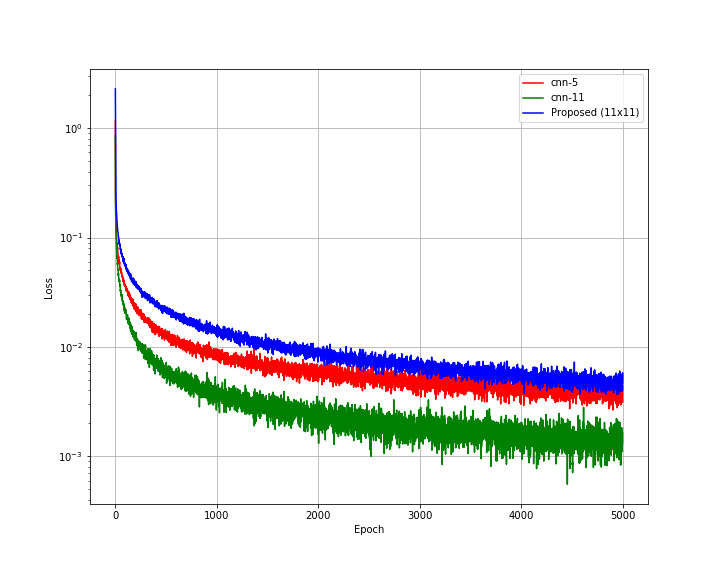}}
	\subfigure[Classification Error.]{\label{fig:b}\includegraphics[width=0.49\linewidth]{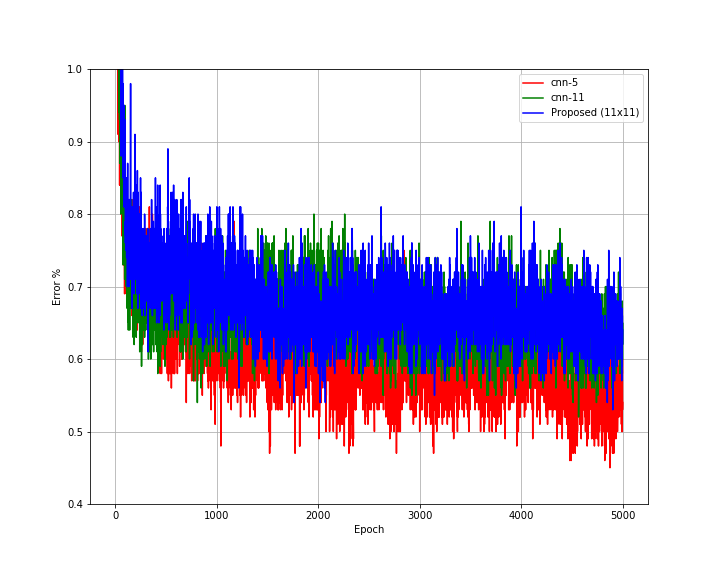}}
	\caption{Training loss and classification error for MNIST.}
	\label{fig:mnistplot}
\end{figure}

In Figure \ref{fig:mnistfilt1}, we can observe the learned envelope functions scale and orientation effects on filters. Smoothing effect of the envelope function over the input is also observed in some outputs (\ref{fig:mnist:filter1:conved}).
Figure \ref{fig:mnistplot} shows the training loss and classification error plots. The adaptive filters had no performance gain against conventional cnn-11, cnn-5. 


\subsection{MNIST Cluttered}
\label{mnistclut}
\begin{figure}
	\centering
	\includegraphics[width=0.49\linewidth]{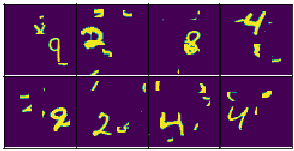}
	\caption{Eight randomly selected samples from the cluttered MNIST dataset.}
	\label{fig:abc}
\end{figure}

Cluttered MNIST dataset \cite{mnistcluttered} consists of 60,000 samples in 10 classes. We split this dataset into 50.000 and 10.000 for test and train purposes, respectively. Randomly selected 8 samples are illustrated in Figure \ref{fig:abc}. It contains 60x60 images generated from the original MNIST database with numerous of distractors. Projecting the original MNIST 28x28 pixel space onto 60x60 also caused changes in scale. Thus, the dataset had scale and rotational variances, in addition to cluttered background noise which makes it a suitable test case to demonstrate the use of adaptive filters.

\begin{figure}
	\centering     
	\subfigure[Training loss, y axis is on log scale.]{\label{fig:a}\includegraphics[width=0.49\linewidth]{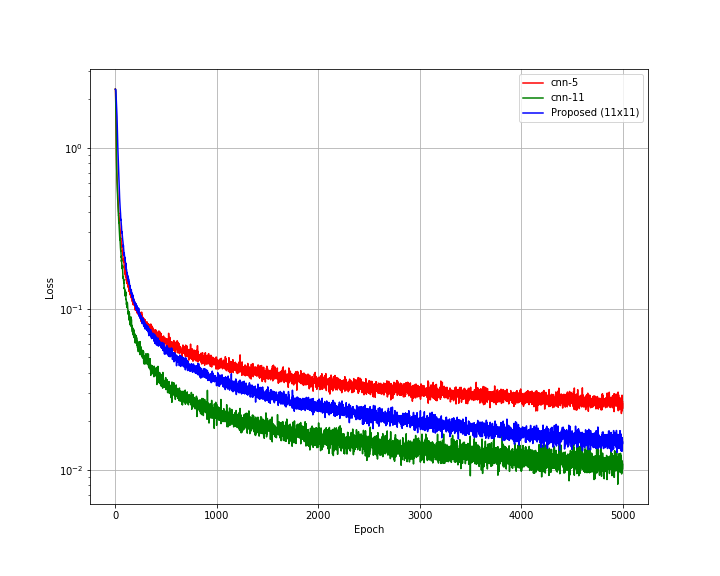}}
	\subfigure[Classification Error.]{\label{fig:b}\includegraphics[width=0.49\linewidth]{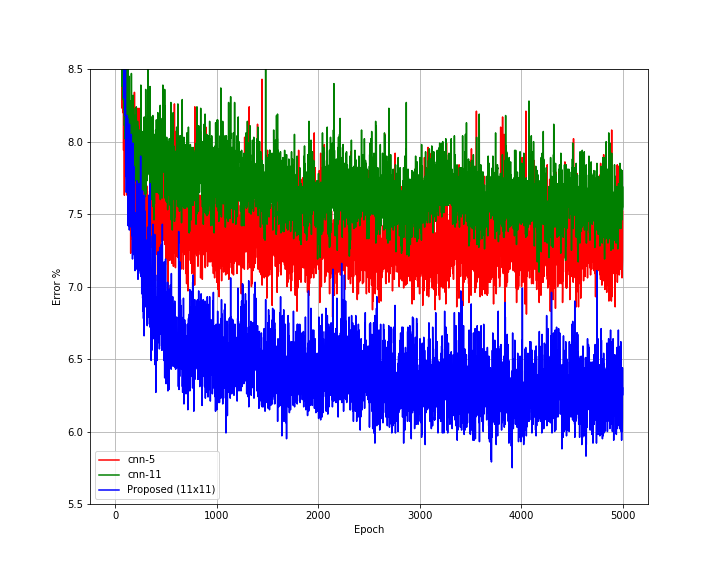}}
	\caption{Training loss and classification error in cluttered MNIST database.}
	\label{fig:mnistclutteredplot}
\end{figure}

Figure \ref{fig:mnistclutteredplot} shows the training loss and classification error, where we can observe better performance compared with conventional CNNs.

\subsection{CIFAR-10}

This dataset \cite{cifar10} is a relatively small (32x32x3) image set with 60,000 samples from 10 different classes. We divided this dataset into 50,000 train and 10,000 test sets, respectively. Other than MNIST dataset, color channels are present, and objects are much more in need for multi-scale features.

\begin{figure}
	\centering
	\subfigure[Gaussian envelope functions.]{\label{fig:a}\includegraphics[width=0.49\linewidth]{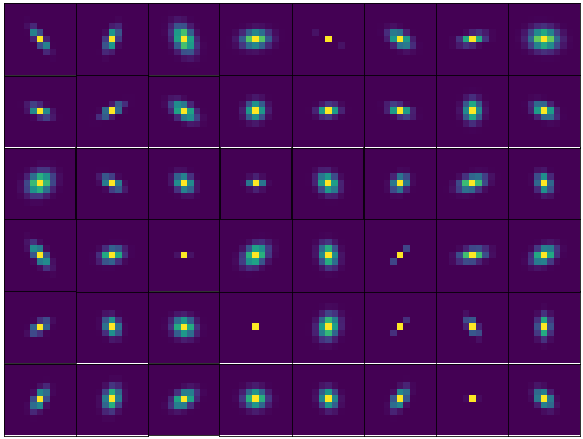}}
	\subfigure[Scaled filters.]{\label{fig:b}\includegraphics[width=0.49\linewidth]{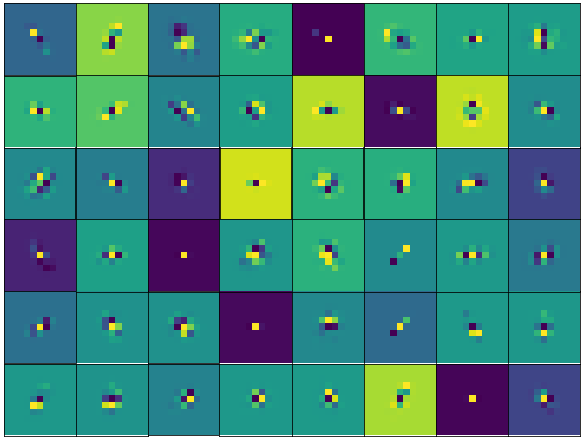}}
	\caption{The first layer filters at the end of training in CIFAR-10 database.}
	\label{fig:cifar10:filter1}
\end{figure}

\begin{figure}
	\centering     
	
	\subfigure[Training loss, y axis is on log scale.]{\label{fig:cifar10:a}\includegraphics[width=0.49\linewidth]{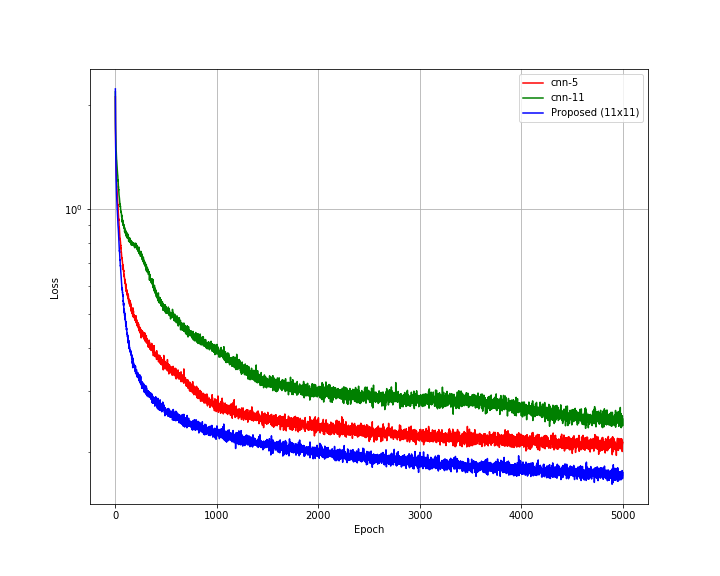}}
	\subfigure[Classification Error.]{\label{fig:cifar10:b}\includegraphics[width=0.49\linewidth]{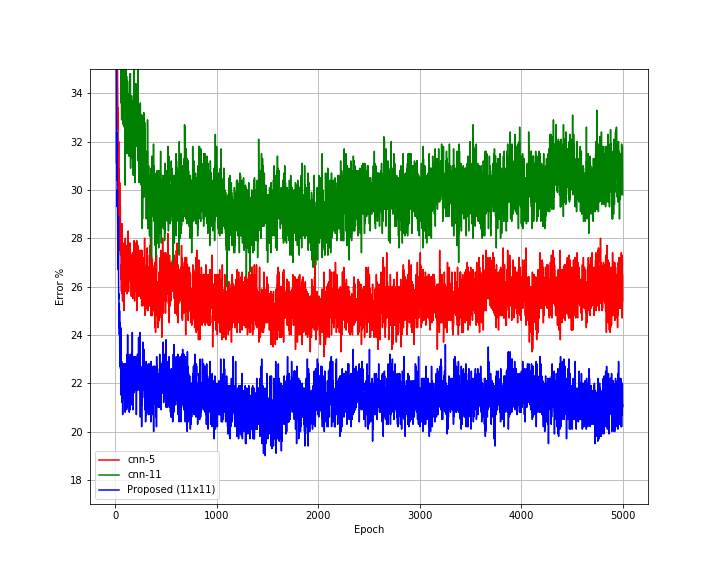}}
	
	\caption{Training loss and classification error in CIFAR-10 database.}
	\label{fig:cifar10:results}
\end{figure}

\begin{figure}
	\centering
	\subfigure[Orientation Change.]{\label{fig:mnist:orientation}\includegraphics[width=0.47\linewidth]{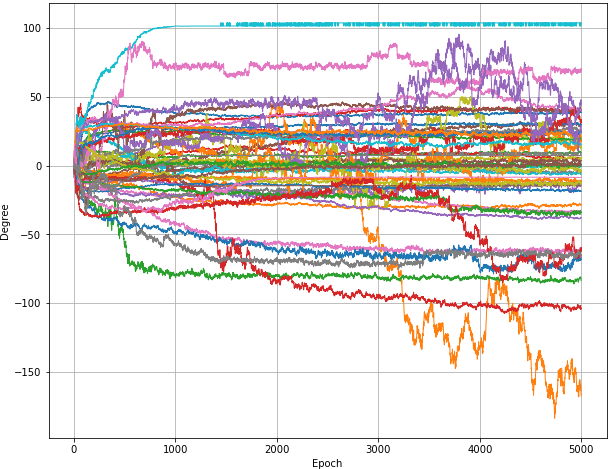}}
	\subfigure[Scale Change.]{\label{fig:mnist:scale}\includegraphics[width=0.47\linewidth]{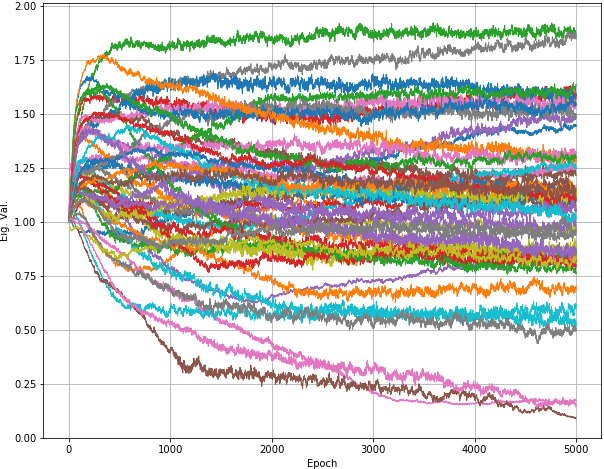}}
	\caption{Plots for covariance matrix $\Sigma$ change in CIFAR-10 dataset, depicted by the (a) angle of largest eigenvector, (b) largest eigenvalue.}
	\label{fig:cifar:convchange}
\end{figure}

The classification results that are shown in Figure \ref{fig:cifar10:b} demonstrates that again ACNN performed better in classification error. For further investigation, we also included the change of $\Sigma$ for learned envelope functions and scaled filters in Figure \ref{fig:cifar10:filter1}. Compared to MNIST, the envelope functions are observably different; and included both large, small, rotated filters. The change in scale and orientation is shown in Figure \ref{fig:cifar:convchange}. Compared with change on $\Sigma$ in MNIST test, scales and orientations have more variation and some of the filters tend to shrink, whereas some were enlarged their scales.

\section{Discussion}

In this paper, we propose an adaptive convolution filter model based on a Gaussian kernel that is acting as an envelope function on shared filter weights. The plots of scale and orientation changes during the training epochs show that the adaptive model is capable of generating differently scaled and oriented filters in a single convolution layer. However, besides bounding and scaling of convolution weights, the Gaussian kernel tends to perform smoothing on input. Such that, if all weights were set to 1 and not trainable, the kernels perform only a Gaussian smoothing operation on input. The initial setting of variance terms to 1.0, enables an initial filter of 5x5 size. During training effective size of the filters are gradually increased. This is because enlarging filters enables more weights to be included in the convolution, which will allow further reduction in the network error. Therefore, the adaptive filter may be prone to overfit more than a conventional fixed sized filter of the same initial size. However, because the envelope rescales weights (max 1.0), it has a regulative effect on their magnitudes, which shall create an advantage. In overall, training of the adaptive filter model did not require very fine tuning of the parameters. However, we observed that the use of dropout layer encouraged the development of filters of different scale and orientation. This can be explained by that the parallel and sparse network configurations induced by the dropout mechanism forces filters to prevent co-adaptation and become independent. We will investigate other ways of inducing independent filters, perhaps with an additional cost term for the network which punishes co-adaptation.

A clear benefit of our model is that it removes the filter size from the list of hyperparameters of deep learning networks. However, our main purpose is to add an adaptive multi-scale representation capacity to convolution layers. The results show that the advantage of using the new model depends on the complexity and variations in training and test data. Among three datasets, MNIST is the simplest where digits are size-normalized and centered \cite{mnist}. The adaptive filters have less or no need for scale adaptation in pixel space, which resulted in no improvement in classification error when compared to conventional CNNs.
However, MNIST cluttered and CIFAR-10 include examples of arbitrary scale, orientation and centering \cite{cifar10} \cite{mnistcluttered}, which allowed the filters to adapt their scale and orientation to improve training while not overfitting. Therefore, we can conclude the adaptive filters expressive power is revealed in datasets with variations in scale and orientation. It is worthwhile to investigate its applications to other domains.

The new and adaptive model of convolution layers allows filters' scale and orientation to be learned during training. Therefore, a single convolution layer can have filters at various scales and orientations. Therefore, a single convolution layer can adapt to extract multi-scale information from its input. State-of-the-art deep networks have many layers and more complex designs compared to the networks that were tested in this study. An interesting question which we will investigate further is whether using the adaptive filter layers can shorten the depth of the state-of-the-art architectures, such as inception \cite{inceptionv1}, highway \cite{highway} or thin \cite{fitnets}. Though our aim is not to fully replace stacked and deep architectures, the new model may help reduce redundancy and improve accuracy. Another question is whether placing the adaptive layer in deeper levels of a network can produce additional gains by focusing on the higher level representations.

\ref{discuss the computational complexity} using Gaussian lookup table for

\section{Acknowledgment}
\label{sec:ack}
This research was supported by a grant (undisclosed for anonymity) and NVIDIA Hardware Grant scheme.

\bibliography{egbib}

\begin{thebibliography}{16}
\providecommand{\natexlab}[1]{#1}
\providecommand{\url}[1]{\texttt{#1}}
\expandafter\ifx\csname urlstyle\endcsname\relax
  \providecommand{\doi}[1]{doi: #1}\else
  \providecommand{\doi}{doi: \begingroup \urlstyle{rm}\Url}\fi

\bibitem[Chan et~al.(2014)Chan, Jia, Gao, Lu, Zeng, and Ma]{pcanet}
Tsung{-}Han Chan, Kui Jia, Shenghua Gao, Jiwen Lu, Zinan Zeng, and Yi~Ma.
\newblock Pcanet: {A} simple deep learning baseline for image classification?
\newblock \emph{CoRR}, abs/1404.3606, 2014.
\newblock URL \url{http://arxiv.org/abs/1404.3606}.

\bibitem[Christopher(2015)]{mnistcluttered}
christopher5106. Christopher.
\newblock Cluttered mnist dataset.
\newblock \url{https://github.com/christopher5106/mnist-cluttered}, 2015.

\bibitem[Dalal and Triggs(2005)]{dalal2005histograms}
Navneet Dalal and Bill Triggs.
\newblock Histograms of oriented gradients for human detection.
\newblock In \emph{Computer Vision and Pattern Recognition, 2005. CVPR 2005.
  IEEE Computer Society Conference on}, volume~1, pages 886--893. IEEE, 2005.

\bibitem[Hubel and Wiesel(1968)]{hubel1968receptive}
David~H Hubel and Torsten~N Wiesel.
\newblock Receptive fields and functional architecture of monkey striate
  cortex.
\newblock \emph{The Journal of physiology}, 195\penalty0 (1):\penalty0
  215--243, 1968.

\bibitem[Krizhevsky and Hinton(2009)]{cifar10}
A.~Krizhevsky and G.~Hinton.
\newblock Learning multiple layers of features from tiny images.
\newblock \emph{Master's thesis, Department of Computer Science, University of
  Toronto}, 2009.

\bibitem[Lecun et~al.(1998)Lecun, Bottou, Bengio, and Haffner]{lenet}
Y.~Lecun, L.~Bottou, Y.~Bengio, and P.~Haffner.
\newblock Gradient-based learning applied to document recognition.
\newblock \emph{Proceedings of the IEEE}, 86\penalty0 (11):\penalty0
  2278--2324, Nov 1998.
\newblock ISSN 0018-9219.
\newblock \doi{10.1109/5.726791}.

\bibitem[LeCun and Cortes(2010)]{mnist}
Yann LeCun and Corinna Cortes.
\newblock {MNIST} handwritten digit database.
\newblock 2010.
\newblock URL \url{http://yann.lecun.com/exdb/mnist/}.

\bibitem[Lindeberg(1994)]{Lindeberg:1994:STC:528688}
Tony Lindeberg.
\newblock \emph{Scale-Space Theory in Computer Vision}.
\newblock Kluwer Academic Publishers, Norwell, MA, USA, 1994.
\newblock ISBN 0792394186.

\bibitem[Lowe(1999)]{lowe}
David~G Lowe.
\newblock Object recognition from local scale-invariant features.
\newblock In \emph{Computer vision, 1999. The proceedings of the seventh IEEE
  international conference on}, volume~2, pages 1150--1157. Ieee, 1999.

\bibitem[Mallat and Zhong(1992)]{mallat1992characterization}
Stephane Mallat and Sifen Zhong.
\newblock Characterization of signals from multiscale edges.
\newblock \emph{IEEE Transactions on pattern analysis and machine
  intelligence}, 14\penalty0 (7):\penalty0 710--732, 1992.

\bibitem[Romero et~al.(2014)Romero, Ballas, Kahou, Chassang, Gatta, and
  Bengio]{fitnets}
Adriana Romero, Nicolas Ballas, Samira~Ebrahimi Kahou, Antoine Chassang, Carlo
  Gatta, and Yoshua Bengio.
\newblock Fitnets: Hints for thin deep nets.
\newblock \emph{CoRR}, abs/1412.6550, 2014.
\newblock URL \url{http://arxiv.org/abs/1412.6550}.

\bibitem[Serre et~al.(2007)Serre, Wolf, Bileschi, Riesenhuber, and
  Poggio]{serre}
T.~Serre, L.~Wolf, S.~Bileschi, M.~Riesenhuber, and T.~Poggio.
\newblock Robust object recognition with cortex-like mechanisms.
\newblock \emph{IEEE Transactions on Pattern Analysis and Machine
  Intelligence}, 29\penalty0 (3):\penalty0 411--426, March 2007.
\newblock ISSN 0162-8828.
\newblock \doi{10.1109/TPAMI.2007.56}.

\bibitem[Srivastava et~al.(2015)Srivastava, Greff, and Schmidhuber]{highway}
Rupesh~Kumar Srivastava, Klaus Greff, and J{\"{u}}rgen Schmidhuber.
\newblock Highway networks.
\newblock \emph{CoRR}, abs/1505.00387, 2015.
\newblock URL \url{http://arxiv.org/abs/1505.00387}.

\bibitem[Szegedy et~al.(2014)Szegedy, Liu, Jia, Sermanet, Reed, Anguelov,
  Erhan, Vanhoucke, and Rabinovich]{inceptionv1}
Christian Szegedy, Wei Liu, Yangqing Jia, Pierre Sermanet, Scott~E. Reed,
  Dragomir Anguelov, Dumitru Erhan, Vincent Vanhoucke, and Andrew Rabinovich.
\newblock Going deeper with convolutions.
\newblock \emph{CoRR}, abs/1409.4842, 2014.
\newblock URL \url{http://arxiv.org/abs/1409.4842}.

\bibitem[{Theano Development Team}(2016)]{theano}
{Theano Development Team}.
\newblock {Theano: A {Python} framework for fast computation of mathematical
  expressions}.
\newblock \emph{arXiv e-prints}, abs/1605.02688, May 2016.
\newblock URL \url{http://arxiv.org/abs/1605.02688}.

\bibitem[Zeiler and Fergus(2014)]{Zeiler2014}
Matthew~D. Zeiler and Rob Fergus.
\newblock \emph{Visualizing and Understanding Convolutional Networks}, pages
  818--833.
\newblock Springer International Publishing, Cham, 2014.
\newblock ISBN 978-3-319-10590-1.
\newblock \doi{10.1007/978-3-319-10590-1_53}.
\newblock URL \url{http://dx.doi.org/10.1007/978-3-319-10590-1_53}.

\end{thebibliography}
\end{document}